\documentclass{article}

\usepackage{microtype}
\usepackage{graphicx}
\usepackage{subfigure}
\usepackage{booktabs} 

\PassOptionsToPackage{hyphens}{url}\usepackage{hyperref}

\usepackage[preprint]{confr2024}

\usepackage{amsmath}
\usepackage{amssymb}
\usepackage{mathtools}
\usepackage{amsthm}
\usepackage[ruled,algo2e]{algorithm2e}

\usepackage[capitalize,noabbrev]{cleveref}

\usepackage[utf8]{inputenc}
\usepackage[switch]{lineno}
\usepackage{commath}
\usepackage{siunitx}
\usepackage{eurosym}
\usepackage[shortlabels]{enumitem}
\usepackage[footnotehyper]{nicematrix}
\usepackage{xspace}
\usepackage{subcaption}
\usepackage[flushleft]{threeparttable}
\usepackage{multicol}
\usepackage{wrapfig}

\newcommand{\cotwo}[0]{CO$_2$\xspace}
\newcommand{\eddp}[0]{($\epsilon, \delta$)-DP\xspace}
\def\aia{AI Act\xspace}
\DeclareSIUnit{\EUR}{\text{\euro}}
\DeclareSIUnit{\kwh}{\text{kWh}}
\DeclareSIUnit{\gcotwoe}{\text{$\mathrm{gCO}_2 \mathrm{e}$}}
\newcommand{\consumerprice}[0]{\SI{0.29}{\EUR\per\kwh} }

\newcommand{\cotwoemissions}[0]{\SI{252}{\gcotwoe\per\kwh} }
\newcommand{\TTF}[0]{TTA50}

\newcommand{\datagov}[1]{{\color[HTML]{000000} #1}} 
\newcommand{\energyeff}[1]{{\color[HTML]{000000} #1}} 
\newcommand{\privacy}[1]{{\color[HTML]{000000} #1}} 

\makesavenoteenv{tabular}
\makesavenoteenv{table}

\theoremstyle{plain}

\confrtitlerunning{
Federated Learning Priorities Under the EU AI Act}
\begin{document}

\twocolumn[
    \confrtitle{
    Federated Learning Priorities Under \\ the European Union Artificial Intelligence Act
    }

    \begin{confrauthorlist}
        \confrauthor{Herbert Woisetschl\"ager}{tum}
        \confrauthor{Alexander Erben}{tum}
        \confrauthor{Bill Marino}{cam}
        \confrauthor{Shiqiang Wang}{ibm}\\
        \confrauthor{Nicholas D. Lane}{cam,flo}
        \confrauthor{Ruben Mayer}{ubt}
        \confrauthor{Hans-Arno Jacobsen}{uot}
    \end{confrauthorlist}

    \confraffiliation{tum}{School of Computation, Information and Technology, Technical University of Munich, Germany}
    \confraffiliation{cam}{Department of Computer Science and Technology, University of Cambridge, United Kingdom}
    \confraffiliation{ibm}{IBM T.J. Watson Research Center, United States}
    \confraffiliation{flo}{Flower Labs, Germany}
    \confraffiliation{ubt}{Department of Computer Science, University of Bayreuth, Germany}
    \confraffiliation{uot}{Department of Electrical and Computer Engineering, University of Toronto, Canada}

    \confrcorrespondingauthor{Herbert Woisetschläger}{herbert.woisetschlaeger@tum.de}
    \confrcorrespondingauthor{Alexander Erben}{alex.isenko@tum.de}
    \confrcorrespondingauthor{Bill Marino}{wlm27@cam.ac.uk}
    \confrcorrespondingauthor{Shiqiang Wang}{\mbox{wangshiq@us.ibm.com}}
    \confrcorrespondingauthor{Nicholas D. Lane}{ndl32@cam.ac.uk}
    \confrcorrespondingauthor{Ruben Mayer}{ruben.mayer@uni-bayreuth.de}
    \confrcorrespondingauthor{Hans-Arno Jacobsen}{jacobsen@eecg.toronto.edu}


    \vskip 0.3in
]

\printAffiliationsAndNotice{}

\begingroup
\let\clearpage\relax

    \begin{abstract}
        The age of AI regulation is upon us, with the \emph{European Union Artificial Intelligence Act} (\aia) leading the way.
        Our key inquiry is how this will affect \emph{Federated Learning} (FL), whose starting point of prioritizing data privacy while performing ML fundamentally differs from that of centralized learning.
        We believe the \aia and future regulations could be the missing catalyst that pushes FL toward mainstream adoption.
        However, this can only occur if the FL community reprioritizes its research focus.
        In our position paper, we perform a first-of-its-kind interdisciplinary analysis (legal and ML) of the impact the \aia may have on FL and make a series of observations supporting our primary position through quantitative and qualitative analysis.
        We explore data governance issues and the concern for privacy. We establish new challenges regarding performance and energy efficiency within lifecycle monitoring.
        Taken together, our analysis suggests there is a sizable opportunity for FL to become a crucial component of \aia-compliant ML systems and for the new regulation to drive the adoption of FL techniques in general. 
        Most noteworthy are the opportunities to defend against data bias and enhance private and secure computation. 
    \end{abstract}
    
    \section{Introduction}
    \label{sec:intro}
    On December $8^{\mathrm{th}}$, 2023, the European Union (EU) Commission and Parliament found a political agreement on an unprecedented regulatory framework -- the \emph{EU Artificial Intelligence Act} (\aia) \cite{eu_ai_act,eu_ai_act_website}. 
This is the first, but likely one of many regulations that will affect how ML applications are developed, deployed, and maintained.
In order to comply with this new landscape, ML of all kinds will likely need to undergo significant changes.
Our main focus is on what this means for Federated Learning (FL)~\cite{Zhang2021_survey}, a fundamentally different approach to ML that offers unique benefits, such as privacy~\cite{mothukuri2021survey} and access to siloed data, compared to its more centralized counterpart.
FL enables distributed privacy-preserving learning of models between several clients and a server at scale \cite{mcmahan2017, Tian2022} while the training data never leaves the clients, and only the models are communicated.
We believe that the \aia and subsequent regulations could serve as the catalyst to pushing FL towards mainstream adoption.
However, this will require the FL community to shift some of its research priorities.

In this position paper, we perform a first-of-its-kind interdisciplinary analysis (legal and ML) of the \aia and FL (\Cref{sec:aia}).
Based on our methodology that aligns with the priorities set out by the \aia (\Cref{sec:methodology}), we make several key observations in support of our primary position (\Cref{sec:quant_analysis}):

First, FL struggles to cope with the new performance trade-offs highlighted in the \aia.
As a result, there is a need for a reconsideration of FL research priorities to address these issues, particularly in terms of energy efficiency and the computational costs of privacy.
While governance has been a focus for FL in the past, the \aia brings new challenges, such as performance parity with centralized approaches and lifecycle monitoring under privacy-preserving operations.

Second, FL has inherent advantages over centralized approaches with respect to data lineage and the ability to address bias and related concerns through access to siloed data.
However, there are remaining technical hurdles for data management and governance issues.
At the same time, these technical hurdles have been solved in centralized learning due to its lack of concern for data movement and its effects on privacy.
It is currently unclear how to cope with GDPR at scale and how the right to privacy will be expressed in practice.

Our analysis indicates due to AI regulation that FL has a significant opportunity to become even more widely adopted.
If the FL community can redirect their research efforts to address the new priorities highlighted by the \aia, and combine this with the inherent advantages of FL, it could become the go-to approach for building compliant ML systems.
Therefore, we advocate for a large fraction of the energy that will undoubtedly go into revising all forms of ML to align with the societal values encoded in this act to be directed into FL rather than centralized approaches.
This will lead to us more quickly having access to suitable methods for deployment in this new landscape, and we expect the act to be a new driver (along with the long-standing issue of pure privacy) towards the adoption of FL techniques in general. 

\emph{Our contributions:}\vspace{+0.1cm}
\begin{itemize}
    \vspace{-14pt}
    \item \textbf{Requirement analysis for FL based on the \aia}.
    \datagov{We examine the impact of the \aia on FL systems and methods, outlining requirements and linking them to challenges in FL, aiming to align the legal and ML perspectives.
    }

    \item \textbf{Quantitative and qualitative analysis of FL under the \aia}.
    We quantify the costs associated with FL, identify the current inefficiencies, and discuss the potential energy implications. 
    Through our experiments, we introduce the \privacy{privacy-energy trade-off that arises when fine-tuning a large model} \energyeff{ in a practical FL framework while aiming to be compliant with the \aia.}
    Further, we provide a qualitative understanding of the potential of FL under the \aia. 

    \item \textbf{Future outlook on novel research priorities for the FL community.}
    By distilling our results into a list of future research priorities, we aim to provide guidance such that FL can become the go-to choice for applications incorporating governing EU fundamental rights.
\end{itemize}

    \section{The EU Artificial Intelligence Act}
    \label{sec:aia}
     The \aia's latest draft as of January $23^{\mathrm{rd}}$, 2024 is referenced throughout this section \cite{ai_act_leak}.
This first-of-its-kind, comprehensive, legal framework around AI development and application aims ``[...] \textit{to promote} [...] \textit{trustworthy artificial intelligence while ensuring a high level of protection of health, safety, fundamental rights enshrined in the Charter, including} [...] \textit{environmental protection} [...]'' (Rec.~1)\footnote{We explain the difference between an article (Art.) and recital (Rec.) in \Cref{app:eu_recitals}. When not specified otherwise, Rec. and Art. refer to the EU AI Act.}.
While it is not finalized yet and must be implemented as national law in every EU country, it may set the basis for other non-EU jurisdictions to decide their legislation~\cite{wh_executive_order, canada_act}.
The penalties for violations of the obligations outlined in the \aia are currently set at a maximum of €35M or 7\% of the company's worldwide annual turnover,  whichever is higher (Art.~71.1). 
As such, the fines range in similar dimensions as those of the General Data Protection Regulation (Regulation (EU) 2016/679) (``GDPR'') Art.~83.5.

The \aia differentiates in its classification of AI applications within two dimensions: risk-based (Art.~6) and general-purpose AI models (GPAI) (Art.~52).
We specifically cover the risk-based classification and the associated requirements for high-risk systems (Art.~8).
If an application falls under this ``high-risk'' category, it must follow strict robustness and cybersecurity (Art.~15) and data governance guidelines (Art.~10), including compliance with GDPR.
Additionally, high-risk system providers may soon have to follow energy-efficiency standards once those are finalized by EU standardization entities (Art.~40.2).
As it happens, most applications that benefit from federated aspects fall under this category by default, such as medical applications \cite{pfitzner2021federated} or management of critical infrastructure (electricity, water, gas, heating, or road traffic) \cite{wang2021electricity,el2023federated,tun2021federated,liu2020privacy}.

The root cause of most GDPR infringements is data collection and unlawful processing~\cite{gdpr_tracker}.
The \aia recognizes this fact and emphasizes the importance of the GDPR in its legal text, naming ``\textit{data protection by design and default}'' and ``[...] \textit{ensuring compliance} [...] \textit{may include} [...] \textit{the use of technology that permits algorithms to be brought to the data} [...] \textit{without the transmission between parties}'' (Rec.~45a).
This aligns with the Act's broad insistence that ``\textit{right to privacy and to protection of personal data} [...] \textit{be guaranteed throughout the entire lifecycle of the AI system}'' (Rec.~45a). 
Since FL specifically addresses these privacy concerns and removes data movement and direct access by definition, we must now understand how we can leverage the introduction of the \aia to enable its legal compliance.
For FL, the following three aspects of the \aia are relevant to understand. 

\textbf{Data Governance}. The biggest hurdle that the \aia imposes on high-risk FL applications is data governance, which requires strong oversight of the data that is being used for the entire model lifecycle of development, training, and deployment (Art.~10.2).
The practices shall include an ``\textit{examination in view of possible biases that are likely to affect the health and safety of persons} [...]'' and ``\textit{appropriate measures to detect, prevent and mitigate possible biases}'' (Art.~10.2f,fa).
With these requirements, we can foresee a future where data access is necessary to comply with forthcoming rules.
However, this data access is a reason why data providers might be hesitant to participate, as it is currently unclear how privacy preservation will be enacted and if they might be liable under GDPR.

Federated Learning provides another outlook on this issue.
While the training data is not accessible by design with FL and, thus, cannot easily comply with the requirements under Article~10.2, it can ease the angst of data providers as data is processed on a strict ``need-to-know'' basis and will never be moved from the source.
This can be a more promising path forward to create access to data in a privacy-preserving manner, simply due to the number of participants.
Additionally, FL includes an emerging research field that implements different techniques to reach specific privacy guarantees, which we cover in \Cref{ssec:data_governance_quant}.
While data quality and techniques to detect biases are recognized as having a high priority when developing DL applications~\cite{whang2023data}, FL has to close this gap with indirect techniques to comply with Article~10.
It is up to debate if techniques, e.g., that combat non-IID data in a federated setting~\cite{zhao2018federated}, provide adequate robustness guarantees or if additional safeguards will be needed.

As high-risk applications typically involve personal data and are required to conform to the GDPR Article~10, we take a closer look at how FL is meeting the key requirements of the GDPR:

\textit{Security while processing data}. 
GDPR Art.~32.2 calls for strict security guidelines when processing data: ``[...] \textit{the appropriate level of security account shall be taken} [...] \textit{that are presented by processing, in particular} [...] \textit{unauthorized disclosure of, or access to personal data transmitted, stored or otherwise processed}''.
While minimizing the risk of data leakage without any data movement, FL shares the model updates during training, providing an attack vector.
To combat this, threat models and security measures for misuse of data by gradient inversion or membership inference attacks have been explored thoroughly~\cite{Zhang2023_membership, huang2021evaluating, geiping2020}.
FL is also vulnerable to data poisoning attacks whereby attackers corrupt client-side data in an attempt to sabotage the model, which is being combated by comprehensive benchmarking~\cite{han2023fedmlsecurity,zhao2023fedprompt}.
Nevertheless, research on FL security remains a key task, as new attacks could emerge. 

\textit{The right to information}. 
While access to data is minimized in FL by only sharing model updates, the GDPR reserves the right for individuals to request all information a service provider has stored (GDPR Art.~15, GDPR Rec.~63~\&~64).
This also includes how data has been used for learning models, which is already being evaluated as client participation is a key priority in FL systems.
Existing studies on personalized FL have established accuracy variance and client update norm as metrics to evaluate the value add a client generates for an FL system \cite{Tan2023, chen2022pflbench, Fallah2020}.

\textit{The right of clients to revoke their consent at any time}.
With the \aia installing GDPR as the adjacent privacy regulation, clients in FL systems may make a request to delete their data or revoke their consent to use it at any time (\aia~Art.~17;~Art.~7).
This can lead to two consequences: the removal of any user data stored in the FL system and, depending on interpretations, the need to unlearn the client's training progress from the global model.
Removing the data is trivial, as the data lineage guarantees provided by FL guard the data from being moved from the clients.
There are a few approaches to machine unlearning~\cite{10.1145/3603620}, such as the teacher-student framework~\cite{kurmanji2023towards} or amnesic unlearning~\cite{graves2021amnesiac}.
However, both techniques need access to the training data or even the entire training progress with client-level model snapshots that are usually unavailable in a federated setting.
In the specific case of FL, existing works focus on unlearning entire clients and provide a possibility for GDPR compliance \cite{halimi2022} without direct data access.

\energyeff{
\textbf{Energy Efficiency}.
While we focus on high-risk applications, the \aia also promotes the environmentally sustainable development of AI systems regardless of the application.
A voluntary Code of Conduct (CoC) will be drawn up to create clear objectives and key performance indicators (Art.~69) to help set best practices regarding, among others, energy efficiency. 
It is still up to discussion which high-risk requirements will be included in this CoC, but it is clear that the position of the AI Act reflects a fundamental value of the EU, namely, sustainability.
While state-of-the-art data centers are designed to be energy-efficient and capable of running on mostly regenerative energy~\cite{green_google}, edge clients used in FL are powered by the average energy mix at their locations~\cite{yousefpour2023, owid-energy-mix}.
This is echoed by the current trends, which indicate that specialized edge devices can compete with data-center GPUs on sample efficiency (sample-per-Watt)~\cite{woisetschlager2023fledge}, but only when looked at the raw throughput, and not in time-to-accuracy comparing FL to centralized training (cf.~\Cref{sec:quant_analysis}).
As such, we find a natural trade-off between energy efficiency and privacy that has yet to be quantified (cf.~\Cref{ssec:energy_efficiency_quant}).
Although we see promising progress toward quantifying how and where energy is being consumed in FL applications \cite{Mehboob2023, qiu2023}, there are still many fundamental open challenges. For instance, we need to find consensus on how the energy-cost responsibility is being assigned in FL with devices not owned by the training provider and how it will compare to future energy-consumption baselines.
}

\textbf{Robustness and Quality Management}.
Unsurprisingly, high-risk AI systems should have an ``\textit{appropriate level of accuracy, robustness} [...] \textit{and perform consistently}'' (Art.~15.1), and this should be guaranteed by a quality management system that takes ``[...] \textit{systematic actions to be used for the development, quality control, and quality assurance}'' (Art.~17.1c).
While it is in the interest of the AI providers to guarantee specific performance goals when deploying, the development of an AI system could be severely prolonged by the need for training to be as energy-efficient as possible.
One technique to guarantee model robustness is early stopping to avoid overfitting, which tracks the model performance on a validation dataset~\cite{prechelt2002early}.
From the earlier data governance requirements on representative data, the time to validate a model may increase as the validation dataset becomes large (cf.~\Cref{ssec:robustness_quant}).
Combining frequent validation with the need for energy consumption monitoring poses a new optimization problem:
Is it more energy-efficient to keep clients idle while a subset validates, or should the next round start in parallel with a chance of overfitting and wasting the energy?
As FL stands currently, this shifts the focus towards techniques that increase the validation efficiency per data sample, e.g., as done in dataset distillation~\cite{lei2023comprehensive}.
As we anticipate a trade-off between energy efficiency and quality management, this could lead to an increased performance gap between centralized learning and FL.

    \section{Methodology}
    \label{sec:methodology}
    \begin{table*}[!ht]
        \centering
        \caption{The algorithmic costs estimate how well the privacy mechanisms scale. Especially, the server-side communication provides evidence that the cryptographic methods are significantly more expensive than \eddp.}
        \label{tab:algorithmic_cost_of_privacy}
        \resizebox{\textwidth}{!}{
            \begin{threeparttable}
                \begin{tabular}{l|c|ccc|ccc|l}
    \toprule
     Privacy & Pot. \aia & \multicolumn{3}{c|}{Client} & \multicolumn{3}{c|}{Server} & \\
     Technique &  compliant* & Computation & Communication & Space & Computation & Communication & Space & Algorithm \\
     \midrule

     $(\epsilon, \delta)$-DP** & \checkmark        & $\mathcal{O}(d)$***                        & $\mathcal{O}(1)$          & $\mathcal{O}(d)$          & $\mathcal{O}(|K|)$     & $\mathcal{O}(|K|)$       & $\mathcal{O}(|K|)$ & {\small \citet{Andrew2019}} \\
     SMPC & \checkmark                          & $\mathcal{O}(|K|^2 + |K| \times d)$                   & $\mathcal{O}(|K| + d)$      & $\mathcal{O}(|K| + d)$      & $\mathcal{O}(|K|^2 \times d)$   & $\mathcal{O}(|K|^2 + |K| \times d)$   & $\mathcal{O}(|K|^2+d)$ & {\small \citet{Bonawitz2017}} \\
     HEC & Limited                           & $\mathcal{O}(d)$                                   & $\mathcal{O}(d)$           & $\mathcal{O}(d)$           & $\mathcal{O}(|K| \times d)$               & $\mathcal{O}(|K| \times d)$                   & $\mathcal{O}(d)$ & {\small \citet{Jin2023}} \\
     \bottomrule
     \bottomrule
\end{tabular}




                \begin{tablenotes}
                      \small
                      \item * Potential evaluation for future \aia compliance 
                      \item ** $\mathcal{O}(d)$ for computation originates from clipping a model update. When the FL aggregator is running in a secure enclave, we can also clip updates on the server at cost $\mathcal{O}(|K|\times d)$
                      \item *** $d$ is the dimensionality of $w$ 
                      
                \end{tablenotes}
            \end{threeparttable}
        }
        \vspace{-12pt}
\end{table*}

Our analysis in \Cref{sec:aia} has highlighted a series of core challenges pertaining to data governance without direct data access, energy efficiency, robustness, and overall quality management.
This section presents our evaluation criteria and how they align with the \aia. We also introduce the methodologies for our qualitative and quantitative analysis.

\subsection{Evaluation criteria}

\textbf{Data Governance}.
Data governance in the \aia focuses on data bias reduction and strict enforcement of regulatory privacy.
Our qualitative analysis focuses on identifying the potential of FL to mitigate data bias. 
Therefore, we study the effect FL can have on the availability of data such that a broader data basis becomes available for training. 
A broader and potentially continuously evolving training dataset could improve the generalization capability of a model and better account for minority groups \cite{torralba2011unbiased}.
For privacy, we look into the technical capabilities of private and secure computing currently available to FL applications. 
We study whether there is a gap between state-of-the-art technical privacy methods and the regulatory privacy requirements introduced by the \aia and GDPR.

\textbf{Energy Efficiency}.
In centralized DL, we often fine-tune FMs on servers with multiple GPUs and, thus, require very high bandwidth interconnects ($> 200 \mathrm{GB/s}$) between the GPUs either via NVLink or Infiniband \cite{Li2020, Appelhans2017}.
FL only requires low bandwidth interconnects ($< 1 \mathrm{GB/s}$) since communication happens sparingly compared to multi-GPU centralized learning \cite{Xu2021}.
This creates major design differences in the training process and an entirely different cost model. 
In the following, we point out essential components of the cost model for FL.

The \aia indicates that further guidelines around energy efficiency are forthcoming. When it comes to how those guidelines define and measure energy efficiency, we propose using a holistic methodology that accounts for computation and communication. 
Based on such conservative methodology, we can develop comprehensive baselines to compare against.
The total energy consumption $P$ consists of two major components, computational $P_c$ and communication energy $P_t$, i.e., $P = P_c + P_t$. 

$P_c$ can be measured directly on the clients via the real-time power draw with an on-board energy metering module \cite{beutel2020} or deriving the energy consumption based on floating point operations and a client's system specifications \cite{Desislavov2023}.
At the same time, $P_t$ is generally more challenging to measure as multiple network hops are involved. 
Often, the network infrastructure components, such as switches and routers, are owned by multiple parties and are impossible to monitor for a service provider.
However, the bit-wide energy consumption model is available to calculate the cost of transmitting data \cite{Vishwanath2015}. 
The costs are directly tied to the number of parameters of a client update in an FL system \cite{yousefpour2023}. 
As such, we can calculate the total energy consumption of communication as 
\begin{equation}
\label{eq:communication_cost}
    \begin{aligned}
        P_{\mathrm{t}} = E_t \cdot \mathcal{B} =~& (n_\mathrm{as} \cdot E_\mathrm{as} + E_\mathrm{bng} + n_e \cdot E_e \\ 
        &+ n_c \cdot E_c + n_d \cdot E_d) \cdot \mathcal{B}\mathrm{.}
    \end{aligned}
\end{equation}
From a client to a server, the communication network and its total energy consumption $E_t$ is organized as follows: $E_\mathrm{as}$, $E_\mathrm{bng}$, $E_e$, $E_c$, $E_d$ are the per-bit energy consumption of edge ethernet switches, the broadband network gateway (BNG), one or more edge routers $n_e$, one or more core routers $n_c$, and one or more data center Ethernet switches $n_d$, respectively. 
To get the total energy consumption for communication, we multiply $E_t$ with the size of a model update $d$ in bits $b$, $\mathcal{B} = d \cdot b$. Usually, a model parameter has a precision of $b = 32$ bits but can vary based on the specific application \cite{Gupta2015}.
\citet{Jalali2014} present the per-bit energy consumption for at least one device per network hop that can be used as a guideline. 
While it is possible to trace what route a network package takes \cite{traceroute}, it is currently impossible to track the real energy consumption of a data package sent over the network. 
It specifically depends on what device has been used at what point in the communication chain. 
As such, if the \aia requires us to track the \emph{total energy} consumed by a service, we have to develop solutions to track the networking-related energy consumption. We already see promising progress towards holistically accounting for energy efficiency in FL applications \cite{Mehboob2023,qiu2023,Wiesner2023}.

\textbf{Robustness and Quality Management}.
Aside from energy, the \aia also requires FL service providers to provide a robust model with consistently high performance. 
Since FL does not allow immediate data access, we must find indirect ways to evaluate the model quality and ensure robustness against over-time-evolving input data. 
We look into what indirect strategies exist to control model quality and measure the cost of existing solutions.
Further, we study existing secure and private computing methods with regard to their applicability in FL applications under the \aia that holds the FL service provider liable for any robustness or quality management issues.

\subsection{Quantitative Analysis}
We design experiments to quantify those measurable components of changes we have to introduce to FL systems to comply with the \aia.
We use FL to fine-tune a 110M parameter BERT \cite{devlin2018} model to classify emails of the 20 News Group dataset \cite{LANG1995331}. Such a setup can be found in job application pre-screening tools, which are classified as \emph{high-risk applications} under the \aia.
Details on the training pipeline and the exact experimental setup are available in \Cref{app:exp-details}.

The \aia \emph{data governance} regulation requires FL service providers to adhere to GDPR and protect data by design.
With the absence of data movement in FL applications, we have already taken a major step toward private-by-design applications. 
However, existing research demonstrates that there are still open attack vectors \cite{geiping2020}, and closing them comes at a cost.
We aim to understand the trade-off between scaling a system and the cost incurred by introducing private and secure computation methods (\Cref{ssec:data_governance_quant}).

The forthcoming introduction of the \aia \emph{energy efficiency} directives may require us to implement FL applications with sustainable and energy-saving techniques in mind. 
However, the additional duties to account for data governance, robustness, and quality management require us to frequently analyze the FL model, track the energy consumption of the whole system, and ensure privacy throughout the entire application.
As this introduces a computational overhead, we aim to understand exactly where potentials for improved energy efficiency can be found and how to address them (\Cref{ssec:energy_efficiency_quant}).

The \emph{robustness and quality management} requirements introduce the necessity of closely monitoring the FL model while training. 
This is to ensure consistently high performance.
Close monitoring naturally increases the requirement to communicate and validate the FL model.
This incurs additional costs. 
We evaluate the question of how expensive robustness and quality management are in FL applications and how they could be mitigated (\Cref{ssec:robustness_quant}).

\subsection{Qualitative Analysis}
In our qualitative analysis, we focus specifically on the characteristics of FL that are not empirically measurable. 
To do so, we take the perspective of legislators to look at the qualitative potential of FL.
We aim to identify the potential of FL to serve the fundamental rights of privacy and data bias prevention.
Our objective is to evaluate whether FL has the significant potential to become \emph{the} most adopted privacy-preserving ML technique for high-risk applications under the \aia.

Overall, our analysis aims to add to the understanding of the future potential of FL under the \aia and derive research priorities to help with the broad adoption of FL.

    \section{Analysis}
    \label{sec:quant_analysis}
    Our analysis combines quantitative analysis considering data governance, energy efficiency, as well as robustness, and quality management. 
We expand on our empirical results with a qualitative analysis to identify the characteristics of FL under the \aia that cannot be easily measured.
The \textbf{key insight} are highlighted.

\subsection{Data Governance}
\label{ssec:data_governance_quant}
Secure Multi-Party Computation (SMPC), Homomorphic Encryption (HEC), and $(\epsilon, \delta)$-Differential Privacy (\eddp) all provide technical measures to improve data privacy in FL. 
SMPC and HEC are cryptographic methods that rely on key exchange between clients \cite{Bonawitz2017, Jin2023}.
The client model update encryption removes the ability to track a client's individual contribution toward a global model. 
At the same time, aggregation remains possible as SMPC and HEC keep arithmetic properties.

\textbf{A clear strategy to employing the \emph{right} private and secure computation technique in an \aia-compliant FL system is required}.
We find all methods to come at significant costs (\Cref{fig:dp_experiments} and \Cref{tab:algorithmic_cost_of_privacy}). 
While the cryptographic methods keep the original shape of the model updates in an encrypted form, they require extensive communication and, in the worst case, point-to-point communication between clients. 
This creates practical challenges when scaling an FL system \cite{Jin2023}. 
However, this is where \eddp excels \cite{McMahan2017_dp, Andrew2019}. 
Instead of requiring the clients $K$ to establish a joint secure computation regime, \eddp introduces privacy by model parameter perturbation. 
In detail, we perturb and clip each model weight $w^k_{t+1}\in\mathbb{R}^d$ with dimension $d$ of a client $k \in K$ with random noise $\xi$ sampled from a Gaussian distribution $\mathcal{N}(0, \sigma^{2}_{\Delta})$. 
The variance $\sigma^{2}_{\Delta}$ depends on the number of clients per aggregation round and how many clients have exceeded the clipping threshold in the previous training round.
The quantity $z$ scales the noise that is actually applied to local client update $w^k_{t+1}$ and ultimately determines the degree of privacy we achieve under a constrained privacy budget $\epsilon$ and a data leakage risk $\delta$, 
\vspace{-4pt}
\begin{equation}
    \label{eq:dpfedavg}
    w_{t+1} = \frac{1}{|K|}\sum^{|K|}_{k = 1}\left(w^k_{t+1} + z \cdot \xi \right) \mathrm{.}
\end{equation}
\vspace{-12pt}

As can be seen, the perturbation mechanism benefits from increasing the number of clients in an aggregation round. 
Thus, \eddp is particularly useful for scaled systems, while the cryptographic methods can be useful in smaller systems. 
The optimal strategy for choosing the right privacy technique is yet to be found.

\textbf{It is unclear whether \eddp can be compliant with regulatory privacy as enacted by GDPR}.
While we know that an $\epsilon \leq 1$ provides strong guarantees for privacy, the guarantee always depends on $\delta$ \cite{Dwork2013}.
While setting $\delta$ is trivial in centralized learning, as we know the dataset size before training, it is challenging in FL. 
We cannot be certain about how many clients will eventually participate in the training process and how many data points each client contributes. 
As such, we require heuristics to set $\delta$ appropriately for training in a dynamically evolving FL system. 
An effort to evaluate \eddp for regulatory compliance is planned in the United States \cite{wh_executive_order}; the same should be done in the EU, potentially as a joint effort.

\subsection{Energy Efficiency}
\label{ssec:energy_efficiency_quant}

\textbf{Optimizing for energy efficiency must become a priority in FL research}. 
In our experiment with the BERT pipeline, we find FL to be $5\times$ less energy efficient than centralized training when looking at the computational and communication effort to reach $50\%$ accuracy (\TTF)~(\Cref{fig:baseline_experiments}).
Communication accounts for 26\% of the total energy draw. 
With parameter-efficient fine-tuning (PEFT), we save on computational resources and reduce communication costs significantly.\footnote{{It is important to note if we were to use full-model fine-tuning, the power consumption for computing would amount to 0.48 kWh~and~communication~to~196~kWh to reach \TTF~($2100\times$ more than PEFT). We communicate 110M parameters over 1,057 rounds.}} 
As such, efficient methods for computation in large models can also reduce communication.
First works have shown significant potential for PEFT methods to address energy efficiency but still come at extensive warm-up costs \cite{Babakniya2023} that we need to mitigate in the future.
As the EU has introduced the Emission Trading System-2 (ETS-2), \cotwo emissions are capped by the number of total emission certificates available \cite{abrell2023_co2_price}. 
This immediately impacts the electricity price, which will surge as the ETS-2 Market Mechanism comes into effect in 2027. 
The price for \cotwo certificates is expected to rise by as much as $6\times$, from €45 in 2024 to €300 with a market-made price.
Overall, this increases the need for energy efficiency improvements. 

\begin{figure}
    \centering
    \includegraphics[width=0.49\textwidth]{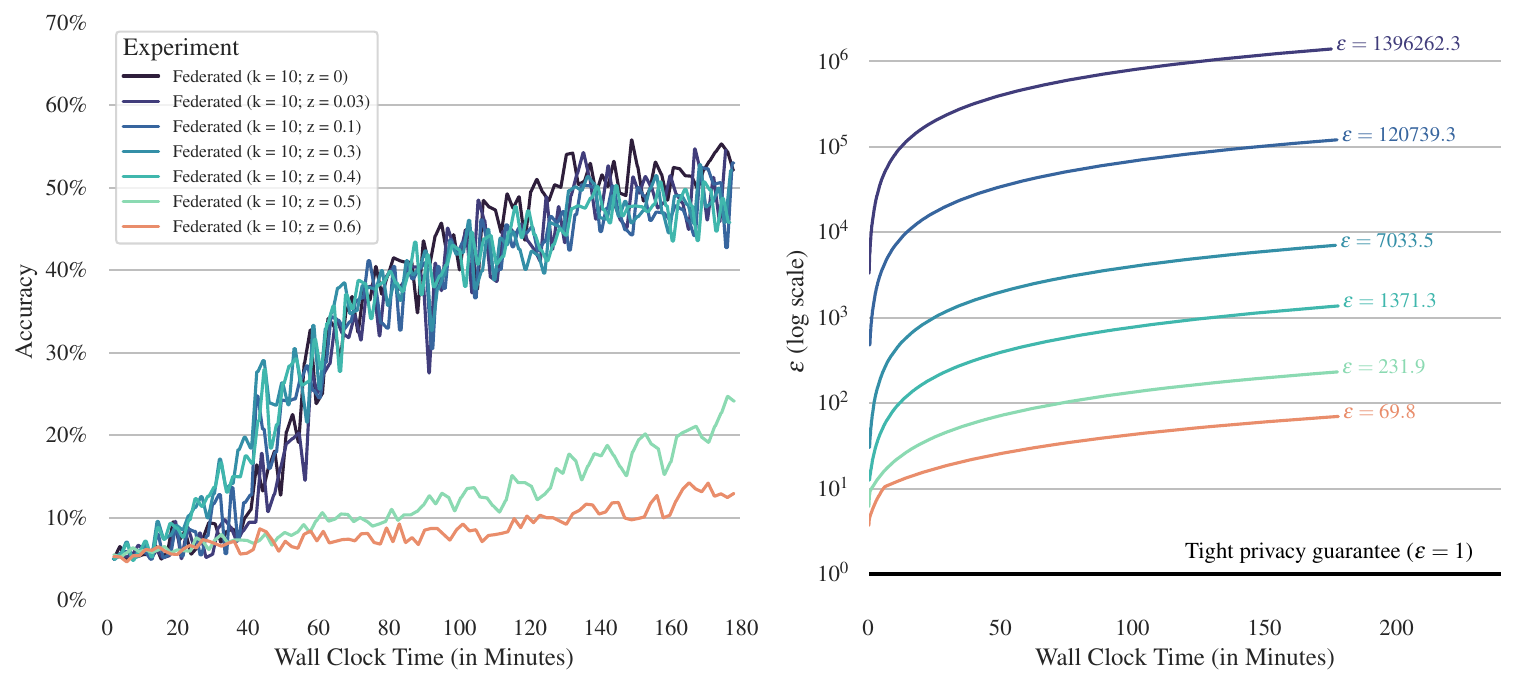}
    \caption{To achieve high privacy guarantees in small systems, we require high $z$ that come at significant model performance and efficiency costs. Training stability also diminishes with increasing $z$. $\epsilon$ is calculated based on $\delta = \frac{1}{16,000}$.}
    \label{fig:dp_experiments}
    \vspace{-16pt}
\end{figure}

\textbf{The \aia introduces a privacy-energy efficiency trade-off}.
As pointed out in \Cref{ssec:data_governance_quant}, we do not know about the \emph{right} choice of private and secure computation for any given FL application as it depends on the number of clients in the system, the number of clients per training round, and the amount of input data available on each client. 
The cryptographic methods introduce significant computational and communication overhead, while \eddp does not. 
However, for small-scale FL systems ($< 100$ clients per aggregation round), the $z$ has to be comparably larger than in large-scale systems \cite{McMahan2017_dp}. 
This significantly reduces the model performance and slows the training process (\Cref{fig:dp_experiments}).
As such, we face a privacy-energy trade-off in current-state FL systems, regardless of the private and secure computation technique.
We must address this challenge in light of the \aia and its call for more energy efficiency.

\subsection{Robustness \& Quality Management}
\label{ssec:robustness_quant}

\textbf{We pay significantly for robustness guarantees}.
Frequent validation in FL under the control of the service provider (Rec. 45), i.e., the server, is a necessity to track model performance, understand a model's robustness against data heterogeneity \cite{li2021fedbn}, and domain shifts \cite{Huang_2023_CVPR}. 
However, the energy consumption of idle clients while waiting for a model to validate and be ready for the next aggregation round has not been part of the power equation thus far.
With the \aia, a service provider may have to account for the \emph{total energy} consumed during training (Art. 40, Rec. 85a).
Thus, we must account for these idle times as well. 
As seen in \Cref{fig:baseline_experiments}, these idle times consume 31\% of all power.
To address this challenge, we could regulate the validation process. 
Similar to what has been done for fair FL methods, we can make validation depend on the loss volatility \cite{Tian2020,Li2020Fair} and validate as follows: 

\vspace{-8pt}
\begin{enumerate}[noitemsep] 
    \item \textit{Only validate the final model}. 
    The fastest way to train is to only validate the final model. 
    However, this approach induces the risk of creating a model with no utility and wasting all energy consumed. Also, legal compliance is in doubt since sparse monitoring contradicts the \aia requirements (Art. 17).

    \item \textit{Validate after every $i^{\mathrm{th}}$ aggregation round}. 
    While a frequent validation strategy reduces the risk of overfitting a model, it creates significant idle time.
    Trading off the validation frequency for energy efficiency could be a promising approach to achieving full compliance with the \aia.

    \item \textit{Validate asynchronously}. 
    We may validate models while starting the next aggregation round to avoid any idle energy consumption.
    This bears the risk of producing an overfitted model but can save energy after all.
    A careful trade-off can help create an energy-efficient system while producing robust models.
\end{enumerate}
\vspace{-8pt}

\textbf{The applicability of HEC under the \aia is potentially limited}. 
Since HEC denies server-side model evaluation by design \cite{Jin2023}, we must rely on client-side validation techniques. 
This is only feasible in applications with trustworthy clients and where validation datasets can be distributed to clients.
Promising directions for trustworthy computing are secure enclaves and trusted execution environments \cite{Sabt2015}.
In the case of client-side validation under the \aia, the FL service provider still remains liable for a consistent and high-quality model. 

\begin{figure}
    \includegraphics[width=0.49\textwidth]{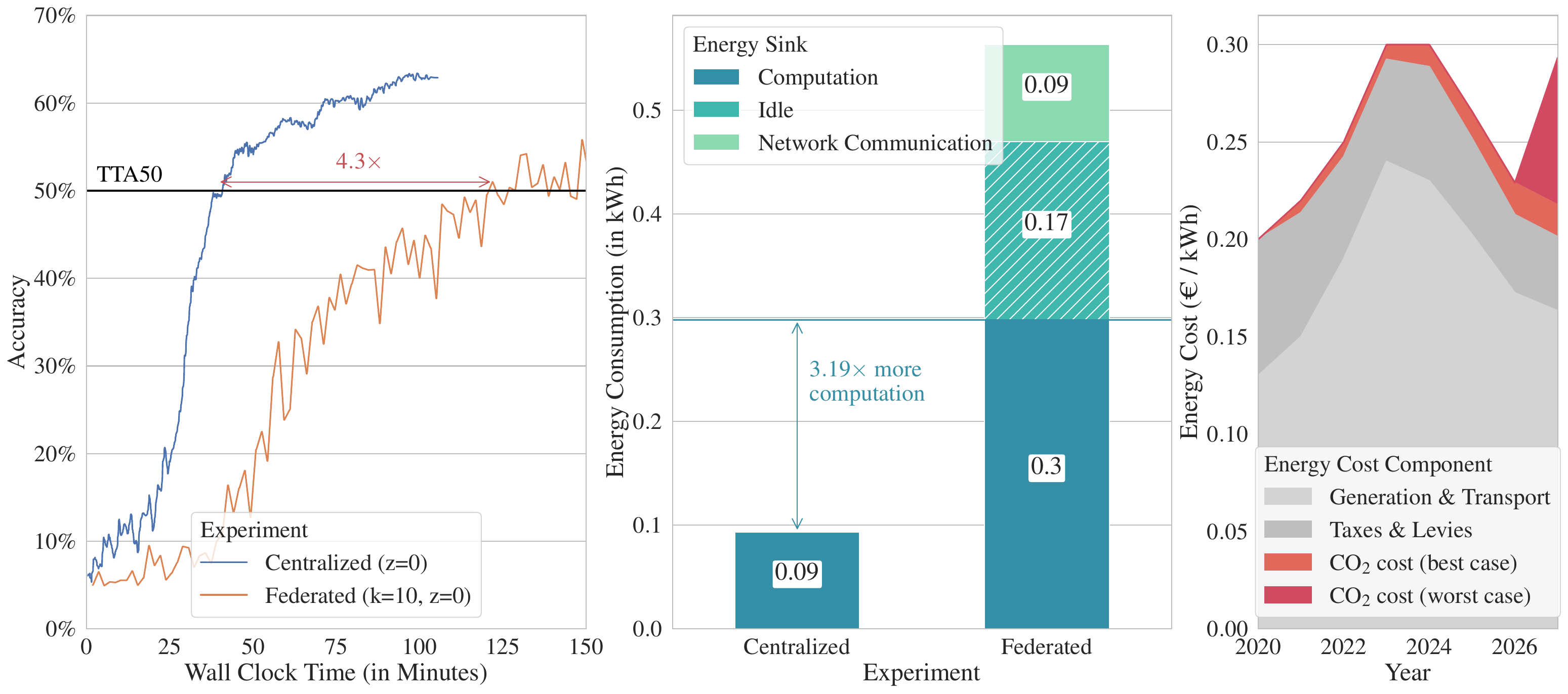}
    \caption{Baseline Experiments. We identify major causes of energy efficiencies in FL systems. The projected energy costs in the EU, especially \cotwo~pricing, require us to focus on improving the energy efficiency of FL.}
    \label{fig:baseline_experiments}
    \vspace{-12pt}
\end{figure}

\subsection{Qualitative Analysis}
\label{ssec:potential_of_fl}

\textbf{Access to siloed data}.
Creating data sets is complex and can, at best, be based on the entire internet~\cite{schuhmann2022laion,gao2020pile}.
Storing and transmitting such huge amounts of data can quickly become costly.
Additionally, data quality is just as important as the data itself~\cite{whang2023data}. 
We assume that a lot of high-quality, simultaneously personally identifiable data is naturally not publicly accessible.
Despite the EU's plan to make anonymized data available worldwide (Rec.~45), collecting such data poses significant challenges, as we outlined in this section.
FL can provide us access to this data, potentially greatly improving the high-risk application's functionality.

\textbf{With broader data access, we generate more representative models and data}.
The \aia emphasizes the importance of examining and mitigating potential biases in the data used for training.
This is particularly important if these biases affect fundamental rights (Art.~10.2f).
To achieve this, the datasets must be curated and prepared for training after they are centrally aggregated.
If a concept shift alters the basic assumptions about the data~\cite{8496795}, the dataset must be adjusted anew.
FL offers a potential solution to this problem.
As FL operates on the clients close to the data source, it means that, by definition, we have access to the latest and most representative data.
Given that we train for many rounds and randomly sample clients for aggregation out of an evolving client base, we automatically create a representative global model over time since the model evolves along with the client base.
As such, it can be easier to comply with the \aia requirements by design.

\textbf{FL provides simple data lineage}.
Since the training data never leaves the clients in FL, it is less complex to track the data lineage, meaning where the data originated, where it has gone, and its usage.
On the one hand, as data is easy to trace, the GDPR requirement to know how the data is being used (cf.~\Cref{sec:aia}) is easy to answer and easier to ensure.
On the other hand, every time data is sent, it is open to man-in-the-middle attacks~\cite{7442758}, and when it is stored in multiple locations, all data hosts are vulnerable to unauthorized access.
Additionally, every time a human is in the loop regarding data management, there is a potential risk of error~\cite{evans2016human}, which can lead to data leakage to a third party.
This lack of vulnerability in FL systems removes most of the potential penalties under the GDPR, which are closing in at €4.5B over the last five years by January 2024~\cite{gdpr_tracker}.
This fact alone can encourage data providers to make data available to FL applications as the risk on their side is significantly lower than before.

    \section{Future Research Priorities}
    \label{sec:future_research}
    The challenges highlighted in our analysis indicate that FL can strongly align with the needs of the \aia if the core challenges are being addressed soon.
To do so, we outline the future research priorities that we see as a necessary redirection for the FL community to make FL a legally compliant and commercially viable solution.
The \textbf{research priorities} are highlighted.

\textbf{The data quality requirements are currently not amenable to FL.}
We need to find solutions to meet the data quality requirements of the \aia under Art.~10 without having direct access to the data.
First, if data quality can be indirectly inferred through techniques with heavy energy investment, how do they compare to direct techniques that require direct data access?
Second, do techniques that combat non-IID data provide enough robustness guarantees to qualify for compliance, and if not, what is missing? 
Third, the \aia mentions that data processing methods at the source are desirable (cf.~\Cref{sec:aia}), but it is not made clear who is responsible for the data if multiple parties are involved.
To make meaningful progress towards the goals of the \aia, it is imperative that the FL research community focuses on improving data quality techniques and ensures that Art.~10, under legal guidance, can be effectively implemented in real-world systems.

\textbf{\cotwo-based optimization to compete with centralized training}. 
FL is currently not achieving competitive energy efficiency compared to centralized training (cf.~\Cref{ssec:energy_efficiency_quant}).
Even if using DP will be considered partially compliant regarding data governance, it results in extensive energy costs, just as training and quality monitoring do.
Therefore, we require new techniques to address these costs concurrently.
While there is ongoing research focused on energy efficiency in specific use cases \cite{yousefpour2023,10041930,10171192,10001569}, there is a need for a designated effort to bridge the gap to centralized baselines, which might be running on fully renewable energy or be more energy efficient by default due to locality.

\textbf{Expression of privacy in the context of the EU AI Act}.
FL is private by design and should fit EU AI Act compliance well.
Unfortunately, we found significant shortfalls in energy efficiency and data governance compared to centralized training (\Cref{sec:quant_analysis}).
If the FL research community does not act now, centralized training may be seen as the best approach for high-risk applications.
This could pose a problem for individuals if privacy is not considered a key component from the outset.
If centralized training is deemed the best approach due to better energy efficiency and easier data governance compliance, it is unclear how the right to privacy will be expressed in practice.
It is crucial that the interpretation of the law, such as with the GDPR and subsequent cookie banners~\cite{eu_cookie_pledge}, does not result in the end-user bearing the entire burden while operators take no responsibility.

\textbf{Privacy-preserving techniques alignment within the \aia}.
We evaluated SMPC, HEC, and \eddp within their current applicability to the energy and data governance aspects and found them to be lacking in multiple ways (cf. \Cref{ssec:data_governance_quant}).
From a technical point of view, we need to work on improving these techniques to be more energy-efficient.
However, researchers should advocate for concrete privacy goals to help align legal and arithmetic privacy.

\textbf{Technical framework for regulatory compliance and representative \aia baselines}.
We require a framework that specifically caters to FL, as it has distinct differences from centralized DL in terms of model lifecycle and data access.
This framework is necessary so that not everyone is faced with complying with the \aia from the outset, but to propose best practices to provide a solid basis (in conjunction with the standardization organizations in Art.~40).
Through this framework, the development of comparable baselines is necessary to set the standard on privacy-by-design deep learning in high-risk applications.
Specifically, this framework should strive to standardize edge hardware comparisons, clarify who is responsible for customer energy costs, and establish clear targets for training and deployment.

    \section{Conclusion}
    \label{sec:conclusion}
    In this position paper, we analyze the \aia and its impact on FL. 
We outline how we need to redirect research priorities with regard to achieving regulatory compliance, the energy-privacy trade-off introduced by the \aia, and the need for new optimization dimensions in FL.
Depending on forthcoming energy efficiency requirements, it may also require us to think about holistic monitoring systems while staying energy efficient.
It is also important to address challenges that have been solved in centralized learning such that FL can keep up.
With this, we, as the FL research community, can send a clear signal to legislation and the broad public that we have a strong interest in making FL \emph{the} distributed privacy-preserving DL technology of the future by incorporating societal priorities into our research.
We can do so by answering the open call by the EU Commission to support the newly established EU AI Office to close the gap between regulatory framing and technical implementation \cite{nature_call_ai_office_2024}.

\endgroup


\section*{Impact Statement}
This paper presents work whose goal is to suggest future research directions that will help ensure that FL, with its worthwhile goal of preserving privacy, aligns with other societal values espoused by the EU AI Act, such as keeping AI systems robust, unbiased, energy efficient, transparent, ethical, and secure, especially for high-risk use cases. This paper transparently addresses the challenges that FL may encounter as regards the data governance, energy efficiency, and robustness provisions of the Act and the associated trade-offs that AI providers must be aware of and responsibly navigate when complying with the Act and the societal ideals it encapsulates.

\section*{Acknowledgements}
We would like to thank Hadrien Pouget at the Carnegie Endowment for International Peace, Yulu Pi at the Leverhulme Center for the Future of Intelligence, Bill Shen at the University of Cambridge, and TheFutureSociety for their help in understanding the legal aspects of the \aia and the White House Executive Order on AI.

This work is partially funded by the Bavarian Ministry of Economic Affairs, Regional Development and Energy (Grant: DIK0446/01) and the German Research Foundation (Grant: 392214008).

We welcome feedback on our work and are open to including aspects that we might have missed in future versions of this paper.

{
    \small
    \bibliographystyle{confr2024}
    \bibliography{main}
}

\newpage

\begingroup
    \appendix
    \onecolumn
    \section*{Appendix}
    \label{sec:appendix}
    
\section{Details on the \aia}
\label{app:eu_recitals}
In this appendix section, we provide additional background on the legal aspects of our work.

\subsection{Article vs. Recitals in the \aia}
In our main paper, we argue with Articles and Recitals. Understanding the difference between both is vital. 
The following explanations are based on \citet{klimas2008law}.

\textbf{Article}. 
An article formulates the actual binding law and defines requirements that need to be implemented in technical solutions. 
This is ultimately what decides on violations.
However, some parts can appear ambiguous and leave room for interpretation.
This is where Recitals come into play.

\textbf{Recitals}. 
They provide interpretation to the Articles and help in guiding what needs to be done to ensure full compliance by reciting elements of the Articles and putting them into context.
As such, Recitals provide procedural details on how to implement a law in practice.
While they form the basis for a common understanding of the \aia, they are not legally binding.

\subsection{The latest \aia version}
By the time of writing this paper late 2023 and early 2024, the official Journal of the European Union hosts the original draft of the \aia, which was released on Apr. $21^{\mathrm{st}}$, 2021. In January 2024, EU policymakers and journalists released the pre-final version of the \aia based on the high public demand. Our work is based on this latest version since it contains the final regulation as it will eventually come into effect. It is available here: \url{https://www.linkedin.com/posts/dr-laura-caroli-0a96a8a_ai-act-consolidated-version-activity-7155181240751374336-B3Ym/} and \url{https://drive.google.com/file/d/1xfN5T8VChK8fSh3wUiYtRVOKIi9oIcAF/view}.

\section{Additional Experimental Details}
\label{app:exp-details}
\begin{table}[!ht]
    \centering
    \caption{Training hyperparameters per training regime.}
    \label{tab:my_label}
    \resizebox{\textwidth}{!}{
        \begin{tabular}{l|l|l|rlrrrrr|rrlrr}
    \toprule
     Training & Data & Tot. Samples & \multicolumn{7}{c|}{Client} & \multicolumn{5}{c}{Server} \\
    regime & Dist. & Seen & MB Size & Optimizer & LR & WD & Mom. & Damp. & Loc. Iter. & K & k & Strategy & LR & Mom. \\
     \midrule
     Centralized     & IID & 80K & 20      & SGD       & 0.01 & 0.001 & 0.9 & 0.9 & 5 & -- & -- & -- & -- & -- \\
     Federated       & non-IID & 80K & 2       & SGD       & 0.01 & 0.001 & 0.0 & 0.0 & 2 & 100 & 10 & FedAvgM & 1.0 & 0.9 \\
     \bottomrule
     \bottomrule
\end{tabular}

    }
\end{table}

Here, we provide additional details about our experimental results.
For our empirical evaluations, we fine-tune the 110M parameter BERT transformer \cite{devlin2018} over the 20 News Group Dataset \cite{LANG1995331} such that we can reliably classify emails into one of 20 categories. For example, such a classification application can be used in a company's human resource processes to screen job applications. Under the AI Act, such a system is considered a high-risk application.

\begin{figure}[!ht]
    \centering
    \includegraphics[width=\textwidth]{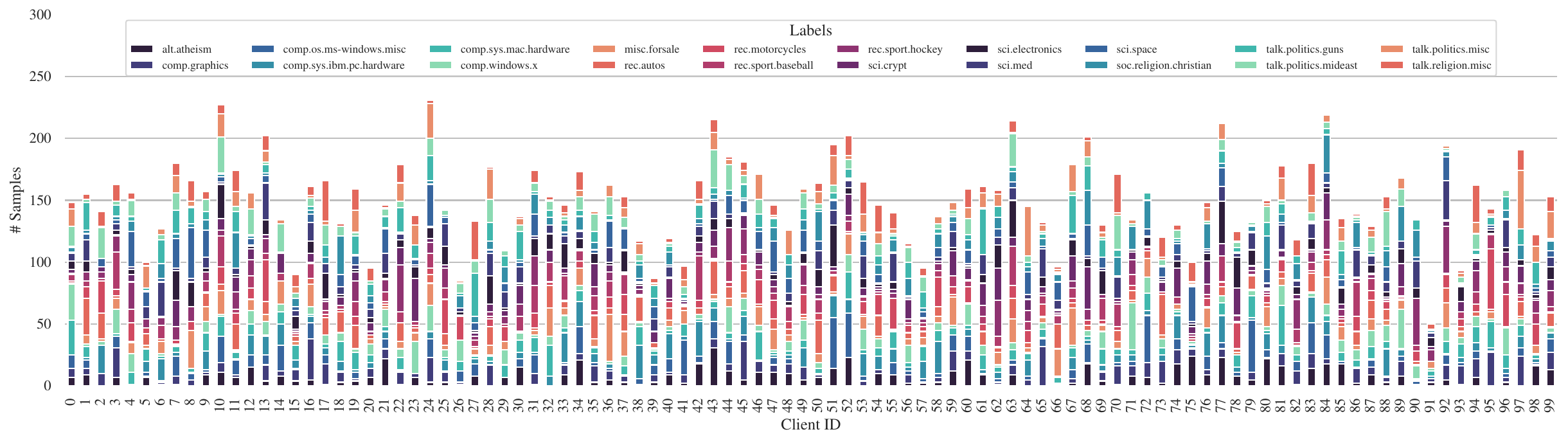}
    \caption{Visualization of client subsets for all of our experiments.}
    \label{fig:subset_distribution}
\end{figure}

\subsection{Dataset}
In our empirical analysis, we use a state-of-the-art text classification task in FL research by means of the 20 Newsgroup Dataset \cite{LANG1995331}, which consists of 18,000 email bodies that each belong to one of 20 classes. 
 The dataset has a total of $18,000$ samples, of which we use $16,000$ for training, $1,000$ for validation, and $1,000$ for testing.
As our work aims to quantify the cost of FL and associated private computing methods in realistic systems in line with the EU AI Act requirements \cite{eu_ai_act}, we chose to sample 100 non-IID client subsets via a Latent Dirichlet Allocation (LDA) with $\alpha = 1.0$, which is widely used in FL research \cite{Babakniya2023, He2020, Reddi2020_adaptive_fl}. The data distribution is visualized in \Cref{fig:subset_distribution}.

\subsection{Model} 
We fine-tune the BERT model \cite{devlin2018} with 110M parameters by using the parameter-efficient fine-tuning technique Low-Rank Adapters (LoRA). 
We use a LoRA configuration that has been well explored in FL settings \cite{Babakniya2023}, which results in 52K trainable parameters (0.05\% of total model parameters). 
This reduces the computational intensity of the task at hand and minimizes the communication load for the FL setup, as we must only communicate the trainable parameters.
The BERT model is used to classify the emails into the 20 distinct categories in the dataset, which resembles a realistic task as it is frequently found in job application pre-screening applications, where the email bodies (input data) often contain sensitive and personal data. 

\textbf{FL configuration}. 
We use the Federated Averaging (FedAvg) algorithm to facilitate all FL experiments \cite{mcmahan2017} and train for $2000$ aggregation rounds. 
We choose a participation rate of $10\%$ for each aggregation round, i.e., $k = 10$ out of $K = 100$.

\textbf{\eddp configuration}. 
We employ sample-level \eddp for centralized learning, and for FL, we use user-level \eddp. Both methods provide the same privacy guarantees \cite{Dwork2013}.
The parameterization for both is identical with $z = [0.0, 0.03, 0.1, 0.3, 0.4, 0.5, 0.6]$ and $\delta = \frac{1}{16,000}$, setting the data leakage risk to the inverse of the number of total training samples \cite{Andrew2019, McMahan2017_dp}. For the experiment with $z = [0.5;0.6]$, we had to change the Learning Rate from 0.01 to 0.001.

\begin{figure}[!ht]
    \centering
    \includegraphics[width=0.45\textwidth]{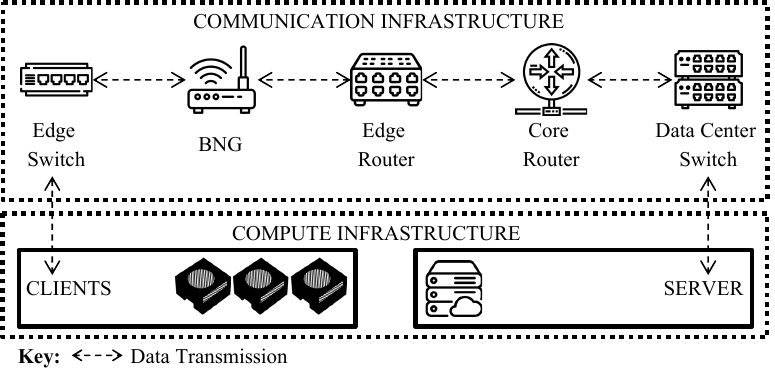}
    \caption{FL system design depicting the network topology for an aggregation round in FL between clients and the aggregation server. Every communication point consumes energy per transmitted bit, which must be accounted for.}
    \label{fig:system-design}
\end{figure}

\textbf{Energy monitoring}. 
We monitor our dedicated clients - NVIDIA Jetson AGX Orin - with 2Hz and measure their total energy consumption while participating in our FL setup. We also use a single Orin device for the centralized experiments for a fair comparison.
For our cost estimations, we use the average price per kWh in the EU, \consumerprice \cite{ConsumerElectricityPrice}. The EU Commission produces quarterly reports on the electricity price trends \cite{energy_market_trends}. Directly proportional to the power consumption, we emit \cotwoemissions \cite{Co2perkWh}.
Regarding communication energy, we assume the average communication route from a private household to a data center with $n_{as} = 1$, $n_{e} = 3$, $n_{c} = 5$, and $n_{d} = 2$ (cf.~\Cref{eq:communication_cost}) \cite{Jalali2014}. For the energy consumption per transmitted bit per network hop, we adopt the values from \citet{Vishwanath2015,Jalali2014} (\Cref{tab:communication_energy_per_hop}).

\begin{table}[!ht]
    \centering
    \caption{Energy consumption per bit network communication for our holistic energy monitoring approach. Values are adopted from \citet{Vishwanath2015,Jalali2014}.}
    \label{tab:communication_energy_per_hop}
    \resizebox{0.7\textwidth}{!}{
        \begin{tabular}{l|l|rr}
    \toprule
     Network Location  & Device Name                    & Upload Cost (nJ/bit) & Download Cost (nJ/bit) \\
    \midrule    
     Edge Switch       & Fast Ethernet Gateway          & 352   & 352 \\
     BNG               & ADSL2+ Gateway (100 Mbit/s)    & 14809 & 2160 \\
     Edge Router       & --                             & 37    & 37 \\
     Core Router       & --                             & 12.6  & 12.6 \\
     Data Center Switch& Ethernet Switch                & 19.6  & 19.6 \\
    
    \bottomrule
    \bottomrule
\end{tabular}
    }
\end{table}

\subsection{Hardware}
\label{app:hardware}
We evaluate the training pipeline on a state-of-the-art embedded computing cluster with NVIDIA Jetson AGX Orin 64 GB devices (Orin), where each device has 12 ARMv8 CPU cores, an integrated GPU with 2048 CUDA cores, and 64 Tensor cores. 
The CPU and GPU share 64 GB of unified memory. 
The network interconnect is 10 GBit/s per client.
We monitor the system metrics with a sampling rate of 2 Hz, including energy consumption in Watt (W).
We use a data center server as an FL server. 
The server has 112 CPU cores, 384 GB of memory, an NVIDIA A40 GPU, and a 40 GBit/s network interface.

\section{Algorithmic Cost Analysis for Private and Secure Computing Techniques in FL}
\label{app:algo_cost}
In this section, we outline how we identified the algorithmic costs of state-of-the-art secure and private computing techniques. We omit the algorithmic costs of FedAvg and focus only on the privacy overhead. We discuss \eddp as introduced by \citet{Andrew2019}, SMPC as introduced by \citet{Bonawitz2017}, and HEC as introduced by \citet{Jin2023}.

\subsection{$(\epsilon, \delta)$-Differential Privacy}
The following algorithm (\Cref{Algo:ada}) is taken verbatim from \citet{Andrew2019}.
For the client, the computational complexity $O(d)$ originates from adding $\xi$ to each parameter of a model update as well as by computing $\Delta$. The communication complexity is $O(1)$ as we need to communicate the standard deviation to parameterize $\xi$ as well as the clipping threshold.
The space complexity $O(d)$ originates from storing $\theta$.

The server computational complexity $O(|K|)$ originates from computing $\tilde{b}^t$ and the communication complexity $O(|K|)$ as we only communicate constants between clients and the server.
The space complexity $O(|K|)$ comes from storing $b_i$.

\begin{algorithm}
    \caption{DPFedAvg-M with adaptive clipping}
    \label{Algo:ada}
    \begin{multicols}{2}
        \begin{algorithmic}
            \FUNCTION{Train($m$, $\gamma$, $\eta_c$, $\eta_s$, $\eta_C$, $z$, $\sigma_b$, $\beta$)}
                \STATE Initialize model $\theta^0$, clipping bound $C^0$
                \STATE $z_\Delta \leftarrow \left( z^{-2} - (2 \sigma_b)^{-2} \right)^{-\frac{1}{2}}$
                \FOR{each round $t=0, 1, 2, \ldots$}
                \STATE $\mathcal{Q}^t \leftarrow$ (sample $m$ users uniformly)
                \FOR{each user $i \in \mathcal{Q}^t$ \textbf{in parallel}}
                \STATE $(\Delta^{t}_i, b^{t}_i) \leftarrow \text{FedAvg}(i, \theta^t, \eta_c, C^t)$
                \ENDFOR
                \STATE $\sigma_{\Delta} \leftarrow z_\Delta C^t$
                \STATE $\tilde{\Delta}^t = \frac{1}{m} \left( \sum_{i \in \mathcal{Q}^t} \Delta^{t}_i + \mathcal{N}(0, I\sigma_{\Delta}^2) \right)$
                \STATE $\bar{\Delta}^t = \beta \bar{\Delta}^{t-1} + \tilde{\Delta}^t$
                \STATE $\theta^{t+1} \leftarrow \theta^t + \eta_s \bar{\Delta}^t$
                \STATE $\tilde{b}^t = \frac{1}{m} \left( \sum_{i\in \mathcal{Q}^t} b_i^t + \mathcal{N}(O, \sigma^2_b) \right)$
                \STATE $C^{t+1} \leftarrow C^t \cdot \exp{\left( - \eta_C(\tilde{b}^t - \gamma)\right)}$
                \ENDFOR
            \ENDFUNCTION
            
            \vspace{7em}
            
            \FUNCTION{FedAvg($i$, $\theta^0$, $\eta$, $C$)}
                \STATE $\theta \leftarrow \theta^0$
                \STATE $\mathcal{G} \leftarrow $ (user $i$'s local data split into batches)
                \FOR{batch $g \in \mathcal{G}$}
                \STATE $\theta \leftarrow \theta - \eta \nabla \ell(\theta;g)$
                \ENDFOR
                \STATE $\Delta \leftarrow \theta - \theta^0$
                \STATE $b \leftarrow \mathbb{I}_{||\Delta|| \leq C}$
                \STATE $\Delta' \leftarrow \Delta \cdot \min{\left(1, \frac{C}{||\Delta||}\right)}$
                \STATE \textbf{return} $(\Delta', b)$
            \ENDFUNCTION

            \vspace{19em}
        
        \end{algorithmic}
    \end{multicols}
\end{algorithm}

\begin{wraptable}{c}{5cm}
  \centering
  \captionof{table}{SecAgg costs}
  \label{table:costs}
  \begin{tabular}{ll}
  \hline
  \multicolumn{2}{l}{\textbf{computation}} \\
  User & $O(|K|^2 + d \cdot |K|)$ \\
  Server & $O(d \cdot |K|^2)$\\
  \multicolumn{2}{l}{\textbf{communication}} \\
  User & $O(|K| + d)$ \\
  Server & $O(|K|^2 + d \cdot |K|)$ \\
  \multicolumn{2}{l}{\textbf{storage}} \\
  User & $O(|K| + d)$ \\
  Server & $O(|K|^2 + d)$ \\
  \hline
  \end{tabular}
\end{wraptable}

\newpage
\subsection{Secure Multi-Party Computation}
The SecAgg algorithmic costs (\Cref{table:costs}) are taken from \citet{Bonawitz2017} Table 1. The naming convention has been adapted to our paper.

\subsection{Homomorphic Encryption}
The following algorithm (\Cref{Algo:hec}) is taken verbatim from \citet{Jin2023}.
For the client, computational complexity $O(d)$ originates from encrypting and decrypting the model. The communication complexity $O(d)$ comes from communicating the aggregation mask once.
The space complexity $O(d)$ is created by storing the aggregation mask.

The server computational complexity $O(|K| \times d)$ originates from the server-side model aggregation while the communication complexity $O(|K| \times d)$ comes from sending the encryption mask once. Storing the encryption mask on the server results in space complexity $O(d)$.

\begin{algorithm}\small
    \caption{HE-Based Federated Aggregation}
    \label{Algo:hec}
    \SetKwFor{ForPar}{for}{do in parallel}{end forpar}
        \begin{itemize}[itemsep=0em]
            \item $[\![\mathbf{W}]\!]$: the fully encrypted model $|$ $[\mathbf{W}]$: the partially encrypted model;
            \item $p$: the ratio of parameters for selective encryption;
            \item $b$: (optional) differential privacy parameter.
        \end{itemize}
    \tcp{Key Authority Generate Key}
    $(pk, sk) \gets HE.KeyGen(\lambda)$;
    
    \tcp{Local Sensitivity Map Calculation}
    \ForPar{each client $i \in [N]$}{
        $\mathbf{W}_i \gets Init(\mathbf{W})$;
    
        $\mathbf{S}_i \gets Sensitivity(\mathbf{W}, \mathcal{D}_i)$;
        
        $[\![\mathbf{S}_i]\!] \gets Enc(pk, \mathbf{S}_i)$;
        
        Send $[\![\mathbf{S}_i]\!]$ to server;
    }
    \tcp{Server Encryption Mask Aggregation}
    $[\![\mathbf{M}]\!] \gets Select(\sum_{i=1}^N \alpha_i [\![\mathbf{S}_i]\!], p$);
    
    \tcp{Training}
    \raggedright
    \For{$t = 1, 2, \dots, T$}{
        \ForPar{each client $i \in [N]$}{
            \If{$t = 1$}{
                 Receive $[\![\mathbf{M}]\!]$ from server;\\
                $\mathbf{M} \gets HE.Dec(sk,  [\![\mathbf{M}]\!])$;\\
            }
            \If{$t > 1$}{
                Receive $[\mathbf{W}_\text{glob}]$ from server;\\
                $\mathbf{W}_i \gets HE.Dec(sk, \mathbf{M} \odot [\mathbf{W}_\text{glob}]) + (\mathbf{1}-\mathbf{M})\odot [\mathbf{W}_\text{glob}]$;\\
            }
            $\mathbf{W}_i \gets Train(\mathbf{W}_i, \mathcal{D}_i)$;\\
            \tcp{Additional Differential Privacy}
            \If{Add DP}{
            $\mathbf{W}_i \gets \mathbf{W}_i + Noise(b)$;
            }
            $[\mathbf{W}_i] \gets HE.Enc(pk, \mathbf{M} \odot \mathbf{W}_i) + (\mathbf{1}-\mathbf{M})\odot \mathbf{W}_i$;\\
            Send $[\mathbf{W}_i]$ to server $\mathcal{S}$;\\
        }
        \tcp{Server Model Aggregation}
        $[\mathbf{W}_{\text {glob}}] \gets \sum_{i=1}^N \alpha_i [\![\mathbf{M} \odot \mathbf{W}_i]\!] + \sum_{i=1}^N \alpha_i ((\mathbf{1}-\mathbf{M})\odot \mathbf{W}_i)$;\\
    }
\end{algorithm}

\endgroup

\end{document}